\documentclass{article}
\pdfoutput=1
\usepackage[utf8]{inputenc}
\usepackage{url}
\usepackage{hyperref}  
\usepackage{cleveref}
\usepackage{xstring}
\IfSubStr{\pdftexbanner}{2009}{%
\usepackage{siunitx}
}{%
\usepackage[load-configurations = version-1]{siunitx}
}
\usepackage{spconf}
\usepackage{graphicx}
\usepackage{amssymb,amsmath,bm}
\usepackage{textcomp}
\usepackage{xcolor}
\usepackage[font={small},skip=2pt]{caption}
\usepackage[tracking=true]{microtype}

\def\baselinestretch{0.87}

\makeatletter
\renewcommand{\section}{\@startsection
  {section}%
  {1}%
  {}%
  {-0.7\baselineskip}%
  {0.3\baselineskip}%
  {}}%

\renewcommand{\subsection}{\@startsection
  {subsection}%
  {2}%
  {}%
  {-0.7\baselineskip}%
  {0.3\baselineskip}%
  {}}%

\renewcommand{\subsubsection}{\@startsection
  {subsubsection}%
  {3}%
  {}%
  {-0.7\baselineskip}%
  {0.3\baselineskip}%
  {}}%

\makeatother

\title{A Comprehensive Study of Deep Bidirectional LSTM RNNs \\
for Acoustic Modeling in Speech Recognition}

\name{{\em Albert Zeyer,
Patrick Doetsch, Paul Voigtlaender,
Ralf Schlüter, Hermann Ney}}

\address{Human Language Technology and Pattern Recognition,
 Computer Science Department, \\
 RWTH Aachen University, 52062 Aachen, Germany \\
  {\small \tt \{zeyer, doetsch, voigtlaender, schlueter, ney\}@cs.rwth-aachen.de}
}

\begin{document}


\maketitle
\begin{abstract}
Recent experiments show that deep bidirectional long short-term memory (BLSTM) recurrent neural network acoustic models outperform feedforward neural networks for automatic speech recognition (ASR). However, their training requires a lot of tuning and experience. In this work, we provide a comprehensive overview over various BLSTM training aspects and their interplay within ASR, which has been missing so far in the literature. We investigate on different variants of optimization methods, batching, truncated backpropagation, and regularization techniques such as dropout, and we study the effect of size and depth, training models of up to 10 layers. This includes a comparison of computation times vs.\ recognition performance. Furthermore, we introduce a pretraining scheme for LSTMs with layer-wise construction of the network showing good improvements especially for deep networks. The experimental analysis mainly was performed on the Quaero task, with additional results on Switchboard. The best BLSTM model gave a relative improvement in word error rate of over 15\% compared to our best feed-forward baseline on our Quaero 50h task. All experiments were done using RETURNN and RASR, RWTH's extensible training framework for universal recurrent neural networks and ASR toolkit. The training configuration files are publicly available.
%
\end{abstract}
\section{Introduction and related work}



Deep neural networks (DNN) yield state-of-the-art performance in classification in many machine learning tasks \cite{schmidhuber2015deepoverview}.
The class of recurrent neural networks (RNN)
and especially long short-term memory (LSTM) networks \cite{hochreiter1997lstm}
perform very well when dealing with sequence data like speech.

Only recently,
it has been shown that LSTM based acoustic models (AM) outperform FFNNs on
large vocabulary continuous speech recognition
(LVCSR) \cite{sak2014lstm,geiger2014robustlstm}.
The training procedure for LSTMs, esp.\ deep bidirectional LSTMs (BLSTM)
takes a lot of time and effort to tune,
arguably more than for feed-forward networks.
There are many aspects to be considered for training LSTMs which we are exploring in this work,
such as the network topology,
sequence chunking and batch sizes, optimization methods,
regularization,
and our experiments show that there is a huge variance in recognition performance
depending on all the different aspects.
What is missing, is an overview over the effect and interdependencies
of the various approaches.
To the best of our knowledge, currently no overview
like this exists the literature,
and this is presented in this work.
We try to fill this gap by a comprehensive study of various aspects
of training deep BLSTMs
and we provide configuration files for all our experiments
\cite{zeyer2016returnn-lstm-experiments}
for our framework RETURNN \cite{doetsch2016crnn}.
Compared to our best FFNN baseline, we get a relative improvement
in word error rate (WER)
of over 15\%.
We train deep BLSTM networks with up to 10 layers for acoustic modeling
and we discovered
that a pretraining scheme
with a layer-wise construction
can improve
performance for deeper LSTMs.
We are not aware of any previous work which applied pretraining for LSTMs
in ASR.

Hybrid RNN-HMM models were developed in 1994 in \cite{robinson1994rnnhmm}.
An early work for bidirectional RNNs for TIMIT was presented in
\cite{schuster1997bidirectional}
and an early hybrid LSTM-HMM was presented in \cite{graves2013hybrid} for TIMIT.
\cite{sak2014lstm,geiger2014robustlstm,
li2015deeplstm,li2015lstmmaxout,
sainath2015cldnn,chan2015tcdnnblstmdnn,
senior2015cdplstm,zhang2015highwaylstm}
investigate various bidirectional and unidirectional LSTM topologies
with optional projection
in some cases combined with convolutional or feed-forward layers
for acoustic modeling in ASR.
%
%
Variations of the LSTM model were studied in
\cite{chung2014empirical,jozefowicz2015empirical,greff2015lstm,breuel2015benchmarkinglstm},
although we only present the standard LSTM without peephole in this work.


\section{LSTM Model and Implementation}
\label{sec:lstm}

We use the standard LSTM model without peephole connections
\cite{gers2003peephole}.
If not otherwise stated, we use bidirectional LSTMs (BLSTM).
\label{sec:impl}
Our base tool is the RASR speech recognition toolkit 
\cite{rybach2011:rasr,wiesler2014:rasr}.
We use RASR for the feature extraction pipeline and for decoding.
%
We extended RASR with a Python bridge to allow many kinds of interactions
with external tools.
This Python bridge was introduced to be able to use
RETURNN, our Theano-based framework
\cite{doetsch2016crnn,bastien2012theano}
to do the training and forwarding in recognition of our acoustic model.
In RETURNN, we have multiple LSTM implementations
and it supports all the aspects which we discuss in this paper.
One particular LSTM implementation is supported by a custom CUDA kernel
which gives us great speed improvements.
We provide more details about this software in \cite{doetsch2016crnn}
and the config files in \cite{zeyer2016returnn-lstm-experiments}.

\section{Comparisons and Experiments}
\label{sec:comp}

\newcommand{\devset}{\textit{eval10}}
\newcommand{\evalset}{\textit{eval11}}
We use a subset of 50 hours from the
Quaero Broadcast Conversational English Speech database \textit{train11}.
The development \devset{} and evaluation \evalset{} sets
consist of about 3.5 hours of speech each.
The recognition is performed using a 4-gram language model.
Further details about the task can be found in \cite{nussbaum2010:quaero09}.

\subsection{Baseline}
\label{sec:baseline}

We use the common NN-HMM hybrid acoustic model \cite{bourlard1994hybrid}.
All acoustic models were trained frame-wise with the cross entropy criterion
based on a fixed Viterbi alignment.
We do not investigate discriminative sequence training in this study.
The input features are 50-dimensional VTLN-normalized Gammatone
\cite{schluter2007gammatone}.
We don't add any context window nor delta frames
for the LSTM because we expect that the LSTM automatically learns
to use the context.
%
We use a Classification And Regression Tree (CART) with 4501 labels.
We also have special residual phoneme types in our lexicon which are used in transcription for
unknown or unintelligible parts.
We remove all frames which are aligned to such phonemes according to our fixed Viterbi alignment.
This means that we have only 4498 output class labels in our softmax layer
and in recognition, we never hypothesize such phonemes.
Our FFNN baseline with 9x2000 layers and ReLU activation function
with a context-window of 17 frames
yields
15.3\% WER on \devset{} and 20.3\% WER on \evalset{}.


Our minibatch construction is similar to e.\,g.\ 
\cite{li2015deeplstm} and described in detail in \cite{doetsch2016crnn}.
One minibatch consists of $n_{\mbox{\scriptsize chunks}}$ number of chunks from one or more corpus segments.
The chunks are up to $T$ frames long and we select them every $t_{\mbox{\scriptsize step}}$ frames
from the corpus.
Our common settings are
$T = 50$, $t_{\mbox{\scriptsize step}} = 25$, $n_{\mbox{\scriptsize chunks}} = 40$,
i.\,e.\ a minibatch size of $2000$ frames.

Our learning rate is not normalized by
$\sum_i T_i$ or $T \cdot n_{\mbox{\scriptsize chunks}}$ so that the
update step stays the same for every mini batch
independent from $n_{\mbox{\scriptsize chunks}}$ or $T$.
Thus, in our case, the total update scale per epoch stays the same
independent from $n_{\mbox{\scriptsize chunks}}$ or $T$.
Only $t_{\mbox{\scriptsize step}}$ will have an impact on the total update scale.

For all experiments, we train 30 epochs.
We have a small separate cross validation (CV) set
where we measure the frame error rate (FER) and the cross entropy (CE).
With the model from epochs 5, 10, 30, the epoch from the best CV FER,
the epoch from the best CV CE, 
we evaluate on \devset{} and \evalset{}.
In the results in our tables, we select the epoch of the best WER on \devset{}.
We also state the epoch. This can give a hint about
the convergence speed or
whether we overfit later.

Despite the optimization method which might already provide
some kind of implicit learning rate scheduling,
we always also use another explicit learning rate
scheduling method which is often called Newbob \cite{doetsch2016crnn}.
We start with some given initial learning rate
and when the relative improvement on the CV CE is less than 0.01 after an epoch,
we multiply the learning rate with 0.5 for the next epoch.

Our standard optimization method is most often Adam
\cite{kingma2014adam}
with an initial learning rate of $10^{-3}$.
We use gradient clipping of $10$ by default.

\subsection{Number of Layers}
\label{sec:numlayers}

We did several experiments to figure out the optimal number of layers.
In theory, more layers should not hurt
but in practice, they often do because the optimization problem becomes harder.
This could be overcome with clever initializations,
skip connections, highway network like structures \cite{srivastava2015highway,zhang2015highwaylstm}
or deep residual learning
\cite{he2015residual}.
We did some initial experiments also in that direction but
we were not successful so far.
The existing work in that direction is also
mostly for deep FFNNs and not for deep RNNs
except for \cite{zhang2015highwaylstm}
which trains deep highway BLSTMs up to 8 layers.

The results can be seen in \Cref{tab:nlayer2}. 
For this experiment,
the optimum is somewhere between 4 to 6 layers.
In earlier experiments, the optimum was at about 3 to 4 layers.
It seems the more we improve other hyperparameters,
the deeper the optimal network becomes.
With pretraining as in \Cref{sec:pretraining},
we get our overall best result
with 6 layers.

We also included the best CE value
on the train dataset and the CV dataset
in \Cref{tab:nlayer2}.
This gives a hint about the amount of overfitting.
We observe similar results as in \cite{he2015residual},
i.\,e.\ deeper networks should in theory overfit even more
but they do not which is probably due to a harder optimization problem.
It also seems as if the CV CE optimum is slightly deeper
than the WER optimum.
That indicates that sequence discriminative training
will further improve the results.



\begin{table}[th]
\caption{\label{tab:nlayer2} {Comparison of number of layers,
layer size fixed to 500 for each forward and backward direction.
Dropout 0.1 + $L_2$, Adam, $n_{\mbox{\scriptsize chunks}} = 40$,
WER on \devset{},
reported on the best epoch.
Note that the CE values are not necessarily from the same epoch as the WER
but they are the minimum from all epochs.
Also, the train CE is accumulated while training, i.e. with dropout applied.
}}
\centerline{
\setlength\tabcolsep{2pt}
  \begin{tabular}{@{} |c|S[tabformat=3.2]|c|c|c|c|c| @{}}
    \hline
    \#layers & \#params[M] & WER[\%]  & epoch & train CE & CV CE \\
    \hline \hline
    1 & 6.7 & 17.6 & 30 & 1.72 & 1.64 \\ \hline
    2 & 12.7 & 14.6 & 16 & 1.25 & 1.39 \\ \hline
    3 & 18.7 & 14.0 & 30 & 1.17 & 1.32 \\ \hline
    4 & 24.7 & \textbf{13.5} & 15 & \textbf{1.16} & 1.29 \\ \hline
    5 & 30.7 & 13.6 & 30 & 1.17 & \textbf{1.28} \\ \hline
    6 & 36.7 & \textbf{13.5} & 30 & 1.22 & \textbf{1.28} \\ \hline
    7 & 42.7 & 13.8 & 30 & 1.24 & \textbf{1.28} \\ \hline
    8 & 48.7 & 14.2 & 19 & 1.29 & 1.31 \\
    \hline
  \end{tabular}
}
\end{table}

\subsection{Layer Size}

In most experiments, we use a hidden layer size of 500
(i.\,e.\ 500 nodes / memory cells for each the forward and the backward direction).
In \Cref{tab:hiddensize} we compare different layer sizes.
Note that the number of parameters increases quadratically.
We see that the optimum for this experiment is at about 600-700
(for 3 layers with Adadelta at about 700),
however a model with size 500 is much smaller and not so much worse,
so we used that size for most other experiments.

We did not investigate projections in this work.
With a projection size of about 500, other groups report
a layer size of up to 2000 \cite{sak2014lstm}.


\begin{table}[th]
\caption{\label{tab:hiddensize} {Comparison of hidden layer size.
5 layers, dropout $0.1$, $L_2$ $0.01$, Adam, $n_{\mbox{\scriptsize chunks}} = 40$.
WER reported on \devset{}, reported for the best epoch.}}
\centerline{
  \begin{tabular}{@{} |c|S[tabformat=3.2]|c|c| @{}}
    \hline
    layer size & \#params[M] & WER[\%]  & epoch \\
    \hline \hline
    500 & 30.7 & 13.6 & 30 \\ \hline
    600 & 43.1 & \textbf{13.5} & 30 \\ \hline
    700 & 57.6 & \textbf{13.5} & 18 \\ \hline
    800 & 74.1 & 13.6 & 30 \\
    \hline
  \end{tabular}
}
\end{table}

\subsection{Topology: Bidirectional vs.\ Unidirectional}

Our original experiment showed that we get quite a huge WER degradation
with unidirectional LSTM networks compared to BLSTMs,
over 20\% relative, 19.6\% WER for unidirectional vs.\ 15.6\% WER for bidirectional,
see \cite{zeyer2016onlinebidir},
although we did not tune the unidirectional network as much.
Other groups confirm that bidirectional networks perform better than unidirectional ones
\cite{schuster1997bidirectional,graves2005framewise}.

This huge WER degradation led to further research
where we investigated how to use bidirectional RNNs/LSTMs
on a continuous input sequence
to do online recognition.
We showed that this is possible and with some recognition delay,
we can reach the original WER.
These results are described in \cite{zeyer2016onlinebidir}.

\subsection{Batching}
\label{seq:batching}
We investigated the effect of different numbers of chunks $n_{\mbox{\scriptsize chunks}}$,
window time steps $t_{\mbox{\scriptsize step}}$ and window maximum size $T$,
resulting in the overall batch size $T \cdot n_{\mbox{\scriptsize chunks}}$.
All experiment were done with the same initial learning rate.
%
We did many experiments with varying $n_{\mbox{\scriptsize chunks}} \in \{20,\dots,80\}$
and got the best results with $n_{\mbox{\scriptsize chunks}} \approx 40$.
For some experiments the performance difference was quite notable better
with $n_{\mbox{\scriptsize chunks}} = 40$ compared to $n_{\mbox{\scriptsize chunks}} = 20$.
This might be because of a better variance and thus more stable
gradient for each minibatch.
Note that a higher $n_{\mbox{\scriptsize chunks}}$ is usually also faster up to a certain
point because the GPU can work in parallel on every chunk.
We usually use $T = 50$.
We did many experiments with fixed $T - t_{\mbox{\scriptsize step}} = 25$
but we often see a slight degradation when $T \ge 100$.
This might be due to the problem being harder to train
because of the longer backpropagation through time
but maybe we need to tune the learning rate or other parameters
more for longer chunks.
Varying $t_{\mbox{\scriptsize step}}$ did not make much difference except that
for smaller $t_{\mbox{\scriptsize step}}$, the training time per epoch
naturally becomes longer because we see some of the data more often.

\subsection{Optimization Methods}
\label{seq:optim}
We compare many optimization methods and variations between hyperparameters
and esp.\ also different initial learning rates
in \Cref{tab:optim}.
We compare
stochastic gradient descent (SGD),
SGD with momentum \cite{polyak1964momentum,sutskever2013initmomentum}
where one variant only depends on the last minibatch (\emph{mom})
and another variant depends on the full history (\emph{mom2}),
SGD with Nesterov momentum \cite{nesterov1983method,sutskever2013initmomentum},
mean-normalized SGD (MNSGD) \cite{wiesler2014:mnsgd},
Adadelta \cite{zeiler2012adadelta},
Adagrad \cite{duchi2011adaptive},
Adam and Adamax \cite{kingma2014adam},
Adam without the learning rate decay term,
Nadam (Adam with incorporated Nesterov momentum) \cite{dozat2015nadam},
Adam with gradient noise \cite{neelakantan2015gradnoise},
Adam with MNSGD combined,
RMSprop \cite{tieleman2012rmsprop} and
an RMSprop inspired method called SMORMS3 \cite{funk2015smorms3}.
We also tried
Adasecant \cite{gulcehre2014adasecant}
but it did not converge in any of our experiments for this ASR task.
We also test the effect of Newbob.
Note that we only use 3 layers, no $L_2$ and a smaller $n_{\mbox{\scriptsize chunks}}$,
which leads to worse results here compared
to some other sections.

One notable variant was also to use several
model copies $n$ which we update independently and which we merge
together by averaging after some $k$ minibatch updates (\emph{upd-mm-$n$-$k$}).
We vary the amount of model copies and after how much batches we merge.
This is similar to the multi-GPU training behavior
described in \cite{doetsch2016crnn}.
This method yielded the best result in these experiments
but we postpone this for further research.

Overall, Adam was always a good choice.
Standard SGD comes close in some experiments
but converges slower.
Newbob was also important.
Note that Newbob also has some hyperparameters and tuning those
will likely yield further improvements.

\begin{table}[th]
\caption{\label{tab:optim} {
Comparison of different optimization methods.
3 layers, hidden layer size 500, dropout 0.1, $n_{\mbox{\scriptsize chunks}} = 20$.
WER5, WER10, bWER and ep is the
\devset{} WER[\%] of epoch 5, 10, best WER[\%] and
the epoch of the best WER, respectively.
}}
\scalebox{0.95}{\parbox{1.05\linewidth}{%
\centerline{
\setlength\tabcolsep{2pt}
  \begin{tabular}{@{} |c|c|c|c|c|c|c| @{}}
    \hline
method & lr & details & {\footnotesize WER5} & {\footnotesize WER10} & {\footnotesize bWER} & ep \\
    \hline \hline
SGD & $10^{-3}$ & - & 17.0 & 16.1 & 15.8 & 30 \\ \cline{2-7}
{} & $10^{-4}$ & - & 17.9 & 15.8 & 14.9 & 26 \\ \cline{3-7}
{} & {} & mom 0.9 & 17.4 & 15.9 & 14.8 & 28 \\ \cline{3-7}
{} & {} & mom2 0.9 & 16.7 & 16.3 & 15.9 & 19 \\ \cline{3-7}
{} & {} & mom2 0.5 & 17.2 & 16.0 & 15.0 & 30 \\ \cline{3-7}
{} & {} & Nesterov 0.9 & 16.9 & 16.1 & 15.8 & 16 \\ \cline{2-7}
{} & $ 0.5 \cdot 10^{-4}$ & - & 19.7 & 17.1 & 15.4 & 30 \\ \cline{3-7}
{} & {} & mom2 0.9 & 16.8 & 15.5 & 15.0 & 30 \\ \cline{2-7}
{} & $10^{-5}$ & - & 32.1 & 22.3 & 18.6 & 30 \\ \cline{3-7}
{} & {} & then lr $10^{-4}$ & 18.7 & 16.2 & 15.0 & 30 \\
\hline
MNSGD & $10^{-4}$ & avg 0.5 & 20.2 & 18.2 & 17.8 & 20 \\ \cline{3-7}
{} & {} & avg 0.995 & 19.1 & 16.8 & 16.4 & 18 \\
\hline
RMSprop & $10^{-3}$ & mom 0.9 & 33.7 & 26.5 & 26.5 & 10 \\
\hline
SMORMS3 & $10^{-3}$ & - & 16.4 & 16.0 & 15.7 & 23 \\ \cline{2-7}
{} & $10^{-3}$ & mom 0.9 & 16.8 & 16.5 & 15.6 & 29 \\
\hline
Adadelta & 0.5 & decay 0.90 & 20.2 & 15.7 & 15.3 & 13 \\ \cline{3-7}
{} & {} & decay 0.95 & 18.4 & 15.7 & 15.1 & 13 \\ \cline{3-7}
{} & {} & decay 0.99 & \multicolumn{4}{c|}{model broken} \\ \cline{2-7}
{} & 0.1 & decay 0.95 & 16.9 & 15.5 & 15.1 & 13 \\ \cline{2-7}
{} & $10^{-2}$ & decay 0.95 & 24.4 & 20.1 & 17.4 & 29 \\
\hline
Adagrad & $10^{-2}$ & - & 16.9 & 16.0 & 15.6 & 29 \\ \cline{2-7}
{} & $10^{-3}$ & - & \multicolumn{4}{c|}{model broken} \\
\hline
Adam & $10^{-2}$ & - & \multicolumn{4}{c|}{model broken} \\ \cline{2-7}
{} & $10^{-3}$ & - & 16.3 & 15.4 & 14.8 & 30 \\ \cline{3-7}
{} & {} & no lr decay & 16.1 & 15.0 & 14.6 & 11 \\ \cline{3-7}
{} & {} & Nadam & 16.1 & 14.8 & 14.7 & 30 \\ \cline{3-7}
{} & {} & grad noise 0.3 & 16.2 & 15.0 & 14.6 & 16 \\ \cline{3-7}
{} & {} & upd-mm-2-2 & 15.8 & 14.9 & 14.5 & 18 \\ \cline{3-7}
{} & {} & upd-mm-3-2 & 15.8 & 14.5 & \textbf{14.3} & 30 \\ \cline{3-7}
{} & {} & Adamax & 16.3 & 15.4 & 14.9 & 15 \\ \cline{2-7}
{} & $0.5 \cdot 10^{-3}$ & - & 15.8 & 14.9 & 14.5 & 13 \\ \cline{2-7}
{} & $10^{-4}$ & - & 16.4 & 15.6 & 14.9 & 18 \\ \cline{3-7}
{} & {} & Adamax & 21.0 & 18.6 & 16.6 & 30 \\ \cline{3-7}
{} & {} & MNSGD & 16.5 & 15.7 & 14.9 & 18 \\ \cline{3-7}
{} & {} & no Newbob & 16.4 & 15.6 & 15.2 & 21 \\ \cline{2-7}
{} & $10^{-5}$ & no Newbob & 30.7 & 24.3 & 19.2 & 30 \\
\hline
  \end{tabular}
}
}}
\end{table}

\label{sec:grad_clip}
We also investigated the effect of various different gradient clipping variants
and we settled with clipping the total gradient for all parameters
with a value of $10$, which stabilized the training in some cases,
although if possible, no clipping yields the best performance in many cases.

\subsection{Regularization Methods}
For regularization, we tried both dropout \cite{hinton2012dropout}
and standard $L_2$ regularization.
The optimal dropout factor
depends on the hidden size and many other aspects,
although we mostly see the optimal WER with dropout 0.1,
i.\,e.\ we drop 10\% of the activations and multiply by $\frac{10}{9}$.
If we enlarge the hidden layer size, we can use higher dropout values
although in most experiments, dropout 0.2 was worse than dropout 0.1.

Interestingly, the combination of both $L_2$ and dropout gives a big improvement and yields
the best result.
See \Cref{tab:regularization}.

\begin{table}[th]
\caption{\label{tab:regularization} {%
We try different combinations of dropout and $L_2$.
3 layers, hidden size 500, $n_{\mbox{\scriptsize chunks}} = 40$, Adam.
WER on \devset{}.}}
\centerline{
  \begin{tabular}{@{} |c|c|c|c| @{}}
    \hline
    dropout & $L_2$ & WER[\%] & epoch \\
    \hline \hline
    0 & 0 & 16.1 & 6 \\ \cline{2-4}
    {} & $10^{-2}$ & 14.8 & 11 \\
    \hline
    0.1 & 0 & 14.8 & 19 \\ \cline{2-4}
    {} & $10^{-3}$ & 14.5 & 11 \\ \cline{2-4}
    {} & $10^{-2}$ & \textbf{14.0} & 30 \\ \cline{2-4}
    {} & $10^{-1}$ & 15.2 & 26 \\
    \hline
  \end{tabular}
}
\end{table}

\subsection{Initialization and Pretraining}
\label{sec:initialization}
\label{sec:pretraining}

In all cases, we randomly initialize the parameters
similar to \cite{glorot2010understanding}.

We investigated the same pretraining scheme as we do for our FFNN
where we start with one layer and add a layer after each epoch
right before the output layer \cite{seide2011feature}.
In each pretrain epoch,
we can either train only the new layer (greedily) or the full network,
where full network training usually was better.

%
Results can be seen in \Cref{tab:pretraining}.
For deeper networks, this scheme seems to help more.
This indicates that our initialization might have room for improvement.
We got our overall best result with such a pretraining scheme
applied for a 6 layer bidirectional LSTM
and we note that esp.\ for the deeper networks,
the improvements by pretraining increases.
We were not able to train a 9 layer BLSTM without pretraining,
it diverged and broke after two epochs.
Also, the training calculation time of the first few epochs is shorter.

\begin{table}[th]
\caption{\label{tab:pretraining} {Comparison of
pretraining and no pretraining,
compared for different final number of layers,
cf.\ \Cref{tab:nlayer2}.
}}
\centerline{
\setlength\tabcolsep{2pt}
  \begin{tabular}{@{} |c|c|c| @{}}
    \hline
    \#layers & \multicolumn{2}{c|}{WER[\%]} \\
    & no pretrain & pretrain \\
    \hline \hline
    1 & 17.6 & - \\ \hline
    2 & 14.6 & 14.4 \\ \hline
    3 & 14.0 & 13.7 \\ \hline
    4 & \textbf{13.5} & 13.4 \\ \hline
    5 & 13.6 & 13.5 \\ \hline
    6 & \textbf{13.5} & \textbf{13.0} \\ \hline
    7 & 13.8 & 13.1 \\ \hline
    8 & 14.2 & 13.3 \\ \hline
    9 & broken & 13.3 \\ \hline
    10 & - & 13.3 \\ \hline
  \end{tabular}
}
\end{table}

\subsection{Calculation Time vs.\ WER}

We did over 300 different training experiments and collected
a lot of statistics about the calculation time in relation to the WER.

Most experiments were done with a GeForce GTX 980.
We see that the Tesla K20c is about 1.38 times slower with a standard deviation of 0.084,
and the GeForce GTX 680 is about 1.86 times slower with a standard deviation of 0.764.
We present the pure train epoch calculation times with a GeForce GTX 980,
not counting the CV test and other epoch preparation.

We collected some of the total times in \Cref{tab:calctimes}.
That is the summed train epoch time until we reach the specific epoch.
We show the model with the best WER up to the specific time.
We see that in most cases, combinations of different
hyperparameters and methods yield the best results.
Time downsampling was a simple method to reduce
the calculation time with performance as trade-off.

\begin{table}[th]
\caption{\label{tab:calctimes} {Total times
until we get to a certain \devset{} WER in a certain train epoch.
If not specified, it's a BLSTM.
}}
\centerline{
\setlength\tabcolsep{3pt}
  \begin{tabular}{@{} |c|c|c|c|c|l|c| @{}}
    \hline
    time & WER & ep & \multicolumn{2}{c|}{model} \\
    {}   & {}  [\%]    & {}    & \multicolumn{1}{c|}{size} & details  \\
    \hline \hline
    2:18h & 20.0 & 5 & 1x500 & dropout + $L_2$ \\ \hline
    2:36h & 17.2 & 5 & 3x300 & dropout + time downsampling \\ \hline
    3:40h & 16.6 & 5 & 3x500 & dropout \\ \hline
    12:30h & 15.3 & 20 & 9x2000 & FFNN with relu \\ \hline
    14:47h & 13.9 & 13 & 3x500 & dropout + $L_2$ \\ \hline
    21:53h & 13.6 & 17 & 4x500 & dropout + $L_2$ + pretraining \\ \hline
    35:36h & 13.2 & 18 & 5x500 & dropout + $L_2$ + grad noise \\ \hline
    41:51h & 13.0 & 23 & 6x500 & dropout + $L_2$ + pretraining \\ \hline
  \end{tabular}
}
\end{table}


\section{Experiments on other Corpora}
\label{sec:swb}
We use the 300h Switchboard-1 Release 2 (LDC97S62) corpus
for training
and the Hub5’00 evaluation data (LDC2002S09) is used for testing.
We use a 4-gram language model which was trained on
the transcripts of the acoustic training data (3M running words)
and the transcripts of the Fisher English corpora (LDC2004T19 \& LDC2005T19)
with 22M running words.
More details can be found in \cite{tuske2015asru}.
Results in \Cref{tab:swb} show the improvement by LSTMs
and also with an associative LSTM \cite{danihelka2016associative}.

\begin{table}[th]
\caption{\label{tab:swb} {Results on Switchboard.
BLSTM models trained with Nadam, gradient noise, dropout + $L_2$.
Additionally with an associative LSTM layer on top.
Cf.\ \Cref{sec:swb}.}}
\centerline{
\setlength\tabcolsep{5pt}
  \begin{tabular}{@{} |c|c|c|c| @{}}
    \hline
    model & \multicolumn{3}{c|}{WER[\%]} \\
    {}    &  total & SWB  & CH \\
    \hline \hline
    FFNN  & 19.4   & 13.1 & 25.6 \\ \hline
    5l.\ BLSTM  & 17.1 & 11.9 & 22.3 \\ \hline
    6l.\ BLSTM  & 16.7 & 11.5 & 21.9 \\ \hline
    5l.\ BLSTM + assoc.\ BLSTM & \textbf{16.3} & \textbf{11.1} & \textbf{21.6} \\ \hline
  \end{tabular}
}
\end{table}

We also did a few experiments on
Babel Javanese full language pack ({\small \texttt{IARPA-babel402b-v1.0b}})
which is a keyword-search (KWS) task
(see \cite{golik15:babel} for all details).
The baseline FFNN with 6 layers and 34M parameters
yields a WER of 54.3\% with CE-training and 53.3\% with MPE-training.
A 3 layer BLSTM with 19M parameters
yields a WER of 52.8\% with CE-training
(without MPE-training yet).

\section{Conclusions}
\label{sec:concl}

%
In this work we studied the effect of various LSTM hyperparameters.
%
We demonstrated how to train deeper LSTM
acoustic models with up to 10 layers.
Important for this achievement
was our introduction of pretraining for LSTMs
which allows for such depth
and is especially relevant for the deeper networks.
We showed that we can
reproduce good results with these findings on several different corpora
and yield very good overall results
which beats our best FFNN on Quaero by over 15\% relatively.
Given our current experience, we think that
(N)Adam is always a good choice for optimization,
some learning rate scheduling like Newbob is important,
and pretraining helps,
esp.\ for deeper models.
Dropout together with $L_2$ regularization works best but should not be too high,
and gradient noise often helps.
First experiments with associative LSTMs were promising.

\section{Acknowledgements}

{
\let\normalsize\footnotesize\normalsize

We thank Zoltán Tüske for the baseline FFNN Switchboard experiment
and Pavel Golik for the Babel baseline experiment.

Partially supported by the Intelligence Advanced Research Projects Activity (IARPA) via Department of Defense U.S.\ Army Research Laboratory (DoD/ARL) contract no.\ W911NF-12-C-0012. The U.S.\ Government is authorized to reproduce and distribute reprints for Governmental purposes notwithstanding any copyright annotation thereon. Disclaimer: The views and conclusions contained herein are those of the authors and should not be interpreted as necessarily representing the official policies or endorsements, either expressed or implied, of IARPA, DoD/ARL, or the U.S.\ Government.

This research was partially supported by Ford Motor Company.
}

\vfill\pagebreak

\bibliographystyle{IEEEbib}
\def\baselinestretch{0.83}
\let\normalsize\footnotesize\normalsize

\let\OLDthebibliography\thebibliography
\renewcommand\thebibliography[1]{
  \OLDthebibliography{#1}
  \setlength{\parskip}{0pt}
  \setlength{\itemsep}{0pt plus 0.3ex}
}

\bibliography{strings,papers.bib}

\end{document}